# Real Time Object Tracking Based on Inter-frame Coding: A Review

[1]**Shraddha Mehta,** [2]**Vaishali Kalariya**
[1,2]RK. University, Rajkot, Gujrat, India

## Abstract
Inter-frame Coding plays significant role for video Compression and Computer Vision. Computer vision systems have been incorporated in many real life applications (e.g. surveillance systems, medical imaging, robot navigation and identity verification systems). Object tracking is a key computer vision topic, which aims at detecting the position of a moving object from a video sequence. The application of Inter-frame Coding for low frame rate video, as well as for low resolution video. Various methods based on Top-down approach just like kernel based or mean shift technique are used to track the object for video, So, Inter-frame Coding algorithms are widely adopted by video coding standards, mainly due to their simplicity and good distortion performance for object tracking.

## I. Introduction
Object tracking is a key computer vision topic, which aims at detecting the position of a moving object from a video sequence. Motion detection was initially developed for video encoding. Motion vectors can be used to predict changes in the scene between two or more video frames. The size of video data is reduced by encoding only the current frame and the motion vectors, from which several future frames can be recovered. Motion detection for object tracking has distinctly different requirements. In object tracking, there is a need to interpret the information given by the motion vectors. In Video compression algorithms all background changes, moving objects, and camera motion need to be encoded, but the meaning of the motion vectors is not important. In Object tracking have to detect only the moving objects and filter out the noise, small camera motions, and insignificant other motions (e.g. rain drops or falling leaves). Grouping or some other kind of association of the motion vectors into a meaningful scene representation is the primary goal of motion detection for object tracking.

## II. Model of Interframe Coding For Object Tracking

### A. Motion Segmentation
The first task is to distinguish the target objects in the video frame. Most pixels in the frame belong to the background and static regions, and suitable algorithms are needed to detect individual targets in the scene.

### B. Noise Removal
Segmentation of the image frame into moving and non-moving pixels, region growing is used to locate and identify each moving object. Each group of pixels that are connected to each other is labeled as one blob. An envelope is established around each object of interest by use of morphological erosion and dilation filters. The erosion-dilation operation also removes blobs that are very small and thus likely to be noise. Additional noise is filtered by ignoring blobs less than S pixels in size.

### C. Representation of a Blob
The variables and their description in the data structure used for representing an individual blob Blob-ID is the number that is given to a blob. The track number of the blob, as assigned by the tracking algorithm, is stored in this variable. The meanR, meanG and meanB variables store the average intensity values of red, green, and blue colors of all pixels belonging to the blob.

### D. Representation of a Track
Track-ID stores the unique track number for the track. Starting-Frame and ending-Frame variables store the first and last frame in which the track was seen, respectively. Lost Count Variable keeps count of how many consecutive frames for which a track has been 'lost'. This is kept so that a track that is lost for more than 5 frames can be deleted and closed.

### E. Object Tracking
It performs object tracking by finding correspondences between tracks in the previous frame and blobs in the current frame.

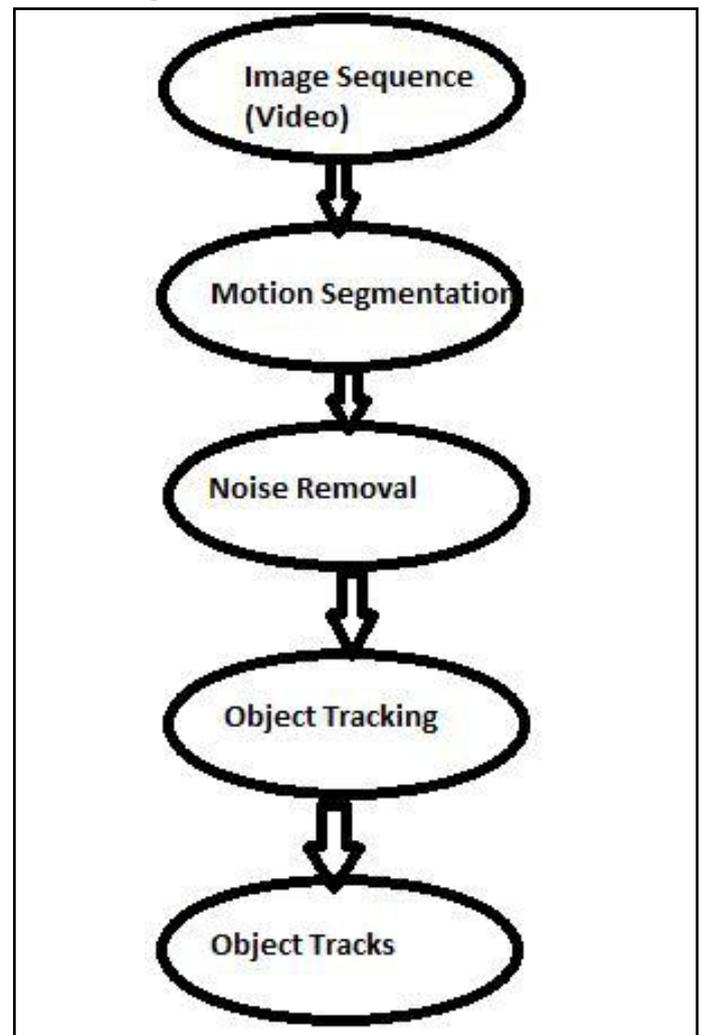

Fig. 1: Model of Inter-frame Coding For Object Tracking

## III. Segmentation
Segmentation defines the divide the image into smaller parts. There is mainly 2 types of segmentation that is Foreground Segmentation and Background Segmentation.

### A. Foreground Segmentation
Detection of object to be tracked is the primary step in object tracking process. The object can be detected either once in the first frame or in every frame in the video. The goal of segmentation is to find out the semantically meaningful regions of an image and cluster the pixels belonging to these regions. It is very expensive to segment all static objects in an image. However, it is more practical to segment only the moving objects from video using spatial-temporal information in sequence of images (frames). A segmentation method should be generic and should not depend on other factors like color, shape and motion. Also, segmentation method should not computationally intensive and should require less memory.

### B. Background Subtraction
The background subtraction method is one of the very simple and promising approaches for extracting moving objects from video sequences. In background subtraction approach, we compare current frame with a reference frame known as background image. A significant difference indicates the presence of moving objects.

## IV. Top- Down Approach
Top down Approach is goal oriented and find the positions of the object in the current frame using a hypothesis generated at the start of the tracking based on parametric representation of the target. The top-down approach uses different block sizes in a manner similar to image pyramids. The basic inter-frame coding is applied using a large block size. The goal is to have the entire object covered only with few blocks. Then the blocks size is reduced by half in each dimension and the block matching is applied again but only for those blocks for which motion was detected during the first step. This can be repeated as many times as desired or until the block size becomes 1*1 pixels.

So, In Short in this approach object has Large Block just like 1*1 pixels then it divided into 2*2 pixel size of blocks and then divided into 4*4 pixel size of blocks in that entire Object is covered.

So, for using the Top Down Approach we can use the formula for n number of levels is : 2n*2n.where n=1,2,3,4…. Depends on the how much the object is large for calculate the motion.

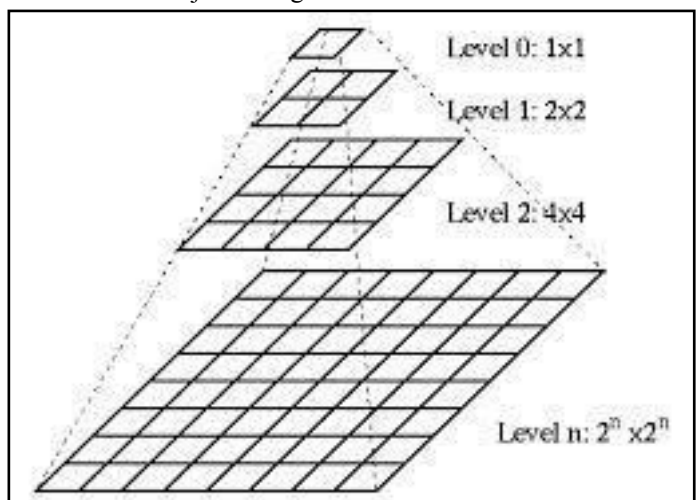

Fig. 2: Top-Down Approach

The top down approaches often rely on external input to initialize the tracking process. These tracking methods use different object characteristics, such as color, texture, shape and motion. In which we only focus on Motion in this paper. One of the popular method in this category is mean-shift based tracking or kernel based tracking.

A regular pyramid is a hierarchy .Each level $fÉ$ contains an array of cells. A cell of a regular pyramid is determined by its position $(i, j, fÉ)$ in the hierarchy, $(i, j)$ are its coordinates within the level $fÉ$. The cells on $fÉ= O$ (base level) are either directly the pixels of the input image or the result of any local computation, like filters, on the image. We obtain each pyramid level recursively by processing the level below.

The reduction window gives us the children-parent relationships. Each cell in level $\lambda +1$ has a reduction window of N×N children at level $\lambda.y$.

## V. Mean-Shift Technique
While mean-shift based technique we use number of points around of one dot detect any possible translation and uniform scaling transformations And one point inside alerts us when we lose the object which detects and estimates the changes in the object movement. We assume a textured object with continuous and smooth motion. We consider translation, rotation and uniform scaling transformations.

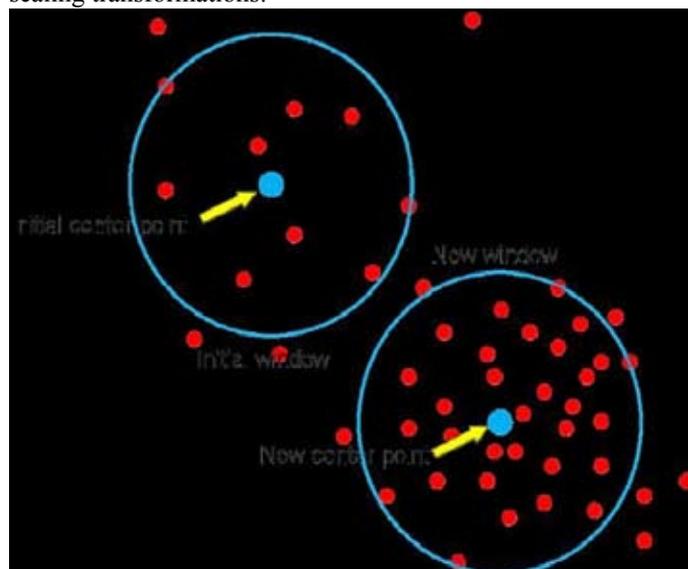

Fig. 3: Mean Shift Technique

## VI. Process of Mean Shift Tracking
- Start from the position of the model in the current frame
- Search in the model's neighborhood in next frame
- Find best candidate by maximizing a similarity function
- Report the same process in the next pair of frames

## VI. Example of Object Tracking
In this Example Object is considered as a man which is moving while the boll is coming in front of him. So in this only man's face is tracked for 58 Seconds. So, this example is useful to understand the object tracking using mean-shift Technique.

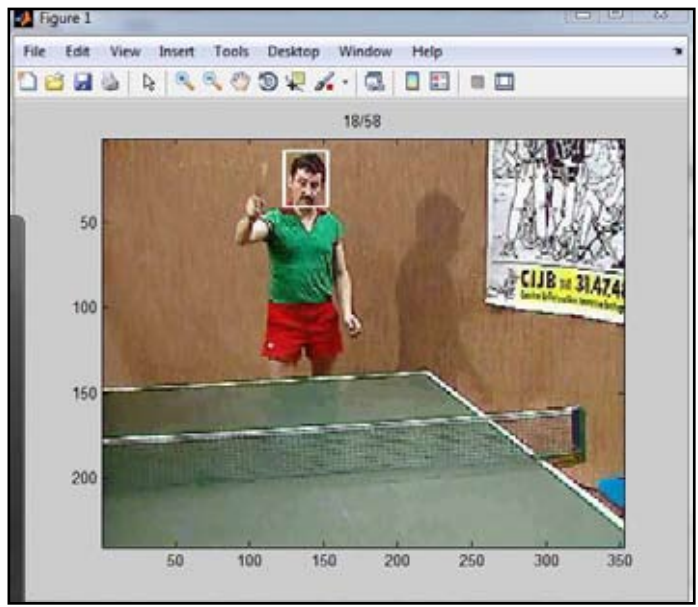

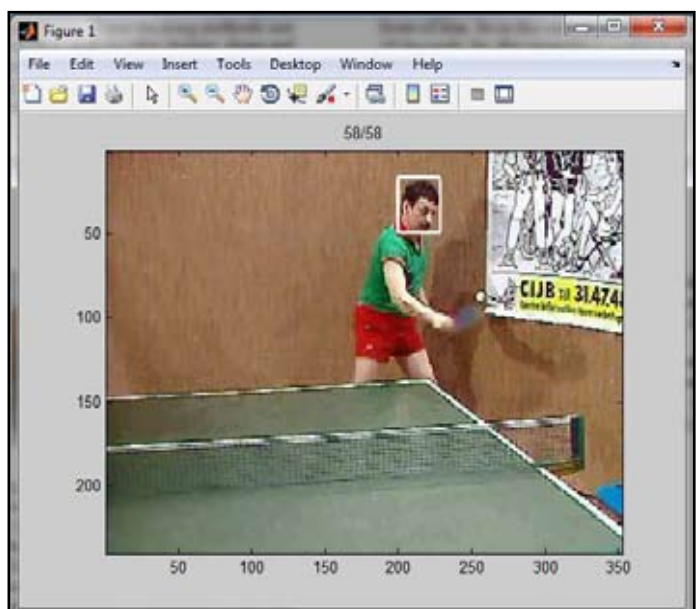

Fig. 4: Tracking Results: Video Sequences - Tracks 18 and 58 Through a Set of Frames

## VII. Result of Mean Shift

Experiment was performed to assess the tracking performance of Mean Shift Tracking. The mean shift based tracker proved to be robust to partial clutter and camera motion. Since no motion model has been assumed, the tracker adapted well to the non-stationary character of the pedestrian's movements. It starts from the position of the model in the first frame and then searches in the model's neighborhood in next frame, followed by finding best candidate by maximizing a similarity function. And then repeats the same process in the next pair of frames.

Table 1: Extracted Features For Tennis Video Sequences

| (Frame, Object) | Area | Width | Height |
|---|---|---|---|
| (1,1) | 250*350 | 6 | 160 |
| (18,1) | 250*350 | 4 | 170 |
| (40,1) | 250*350 | 10 | 190 |
| (47,1) | 250*350 | 15 | 210 |
| (58,1) | 250*350 | 20 | 240 |

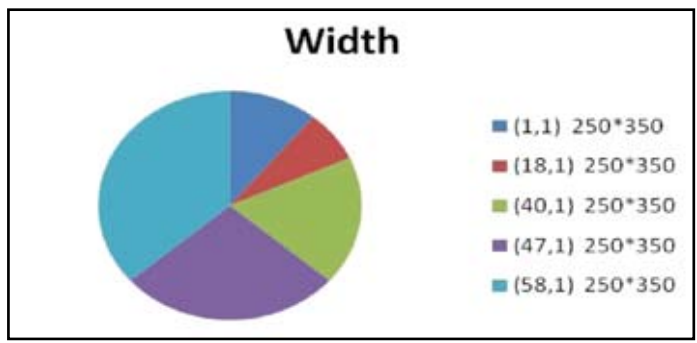

Fig. 5: Visual Tracking Representation

## X. Conclusion

Significant progress has been made in object tracking during the last few years. Several robust trackers have been developed which can track objects in real time in simple scenarios. However, it is clear from the papers reviewed in this survey that the assumptions used to make the tracking problem tractable, for example, smoothness of motion, high contrast with respect to background, etc., are violated in many realistic scenarios and therefore limit a tracker's usefulness in applications like automated surveillance, human computer interaction, video retrieval, traffic monitoring, and vehicle navigation.

## XI. Future Work

While the Texture of Object and Background have Uniform Color Then it will difficult to track that object with motion estimation So, It will be try to solve using top-down approach.

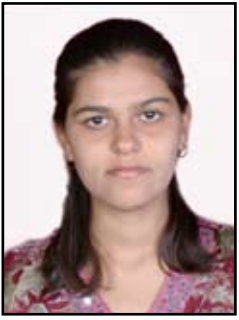

Shraddha Mehta received her B.E. degree in Computer Science Engineering from Gujarat Technological University, India, in 2008, and pursuing M.Tech. degree in Computer Engineering from RK University, Rajkot, India, in 2014, She was a lecturer with Department of Computer Engineering, RK University, in 2012 to 2013.Her research interests includes with Image Processing. At present, She is engaged in Real Time Object Tracking Based On Inter-Frame Coding.

Vaishali Kalariya received her M.E degree in Computer Engineering and pursuing PHD from RK University, Rajkot, India. She is assitant Professor with Department of Computer Engineering, RK University. Her research interests includes with Image Processing.